\documentclass{article}
\usepackage{arxiv}
\usepackage[utf8]{inputenc} 
\usepackage[T1]{fontenc}    

\usepackage{framed}
\usepackage{tcolorbox}
\usepackage{amsmath}
\usepackage{graphicx}
\usepackage[colorlinks=true, allcolors=blue]{hyperref}
\usepackage{array}
\usepackage{ragged2e}
\usepackage{enumitem}
\usepackage{booktabs}
\usepackage{adjustbox}
\usepackage{longtable}
\usepackage{float} 
\usepackage{placeins} 
\usepackage{pgfplots}
\pgfplotsset{compat=1.18}
\usepackage{subcaption}

\definecolor{mygreen}{HTML}{465B3A}
\definecolor{lightgreen}{HTML}{E0E8DC}
\tcbset{
 colback=lightgreen,
 colframe=mygreen,
 boxrule=1mm,
 arc=2mm,
 outer arc=2mm,
 boxsep=3mm,
 left=5mm,
 right=5mm,
 top=3mm,
 bottom=3mm,
 width=\textwidth
}

\newcolumntype{P}[1]{>{\centering\arraybackslash}p{#1}}
\newcolumntype{M}[1]{>{\centering\arraybackslash}m{#1}}

\title{\textbf{Eir: Thai Medical Large Language Models}}
\author{Yutthakorn Thiprak, Rungtam Ngodngamthaweesuk, Songtam Ngodngamtaweesuk, MD \\
    Eir Team
}

\usepackage{tikz}
\usetikzlibrary{shapes.geometric,arrows, positioning, fit, backgrounds}

\tikzstyle{process} = [rectangle, rounded corners, minimum width=1.5cm, minimum height=0.6cm, text centered, draw=black, fill=orange!30, font=\normalsize]
\tikzstyle{data} = [rectangle, minimum width=1.5cm, minimum height=0.6cm, text centered, draw=black, fill=red!30, font=\normalsize]
\tikzstyle{prompt} = [rectangle, rounded corners, minimum width=1.5cm, minimum height=0.6cm, text centered, draw=black, fill=orange!30, font=\normalsize]
\tikzstyle{arrow} = [thick,->,>=stealth]

\tikzstyle{process_cot} = [rectangle, rounded corners, minimum width=1.5cm, minimum height=0.6cm, text centered, draw=black, fill=orange!30, font=\normalsize]
\tikzstyle{db} = [cylinder, shape border rotate=90, minimum height=2cm, minimum width=0.6cm, draw=black, fill=red!30, text centered]
\tikzstyle{llm} = [rectangle, rounded corners, minimum width=1.5cm, minimum height=0.6cm, text centered, draw=black, fill=blue!30, font=\normalsize]
\tikzstyle{cotqa} = [rectangle, rounded corners, minimum width=1.5cm, minimum height=0.6cm, text centered, draw=black, fill=green!30, font=\normalsize]
\tikzstyle{answer} = [rectangle, rounded corners, minimum width=1.5cm, minimum height=0.6cm, text centered, draw=black, fill=green!30, font=\normalsize]
\tikzstyle{arrow_cot} = [thick,->,>=stealth]

\begin{document}
\maketitle
\begin{abstract}
We present Eir-8B, a large language model with 8 billion parameters, specifically designed to enhance the  accuracy of handling medical tasks in the Thai language. This model focuses on providing clear and easy-to-understand answers for both healthcare professionals and patients, thereby improving the efficiency of diagnosis and treatment processes. Human evaluation was conducted to ensure that the model adheres to care standards and provides unbiased answers.

To prioritize data security, the model is deployed within the hospital's internal network, ensuring both high security and faster processing speeds. The internal API connection is secured with encryption and strict authentication measures to prevent data leaks and unauthorized access.

We evaluated several open-source large language models with 8 billion parameters on four medical benchmarks: MedQA, MedMCQA, PubMedQA, and the medical subset of MMLU. The best-performing baselines were used to develop Eir-8B. Our evaluation employed multiple questioning strategies, including zero-shot, few-shot, chain-of-thought reasoning, and ensemble/self-consistency voting methods. Our model outperformed commercially available Thai-language large language models by more than 10\%. In addition, we developed enhanced model testing tailored for clinical use in Thai across 18 clinical tasks, where our model exceeded GPT-4o performance by more than 11\%.

\end{abstract}

\section{Introduction}
Recent advances in artificial intelligence (AI) and large language models (LLMs) have significantly enhanced the capabilities of various natural language processing (NLP) tasks. These advancements present new opportunities to automate functions traditionally performed by humans, such as customer service, language translation, and content generation. Within this context, the potential of AI to assist professionals in various fields, particularly healthcare, is especially noteworthy.

This research aims to explore the application of LLMs in transforming Thailand's healthcare sector by leveraging AI to extract valuable insights from unstructured medical data. These insights are vital for improving population health management, clinical trials, drug discovery, and, ultimately, patient outcomes and healthcare delivery.

LLMs have the ability to identify key data points within electronic health records and digital medical datasets, aiding in the development of new drugs and treatment plans. A notable advantage of LLMs is their capacity for Zero-Shot Learning, which enables them to adapt to new tasks through simple instructions, even without prior exposure. Advanced techniques such as Chain-of-Thought (CoT) \cite{Cot} enhance the model's deep reasoning and decision-making capabilities, resulting in improved accuracy for complex problem-solving tasks. Additionally, Agent-Based Modeling facilitates the efficient management of intricate tasks by distributing subtasks among various subagents \cite{AgentHealthcare}, optimizing data handling.

Despite these advantages, the versatility of LLMs raises significant privacy concerns, particularly in the healthcare domain, where the confidentiality of patient information is paramount. Ensuring robust privacy protections while maintaining the reliability and usability of LLMs remains a critical challenge that must be addressed.

While LLMs have demonstrated success across various domains, their performance in healthcare often falls short due to the lack of domain-specific training. Preliminary findings indicate that the direct application of LLMs to tasks such as biological named entity recognition (NER) and relation extraction (RE) results in suboptimal performance compared to specialized models. Furthermore, integrating LLMs into hospital systems introduces privacy risks, as many LLMs are accessible only via external APIs, making it impractical to upload sensitive patient data.

In Thailand, the development of NLP technology that supports the Thai language is still in its nascent stages. The complexity of Thai grammar, coupled with the diverse idiomatic expressions, presents significant challenges in the creation of effective language models. Additionally, the lack of high-quality Thai language resources remains a persistent issue.

We hypothesize that a robust language model can be effectively adapted to the Thai language through additional training on a moderately sized, Thai medical-specific dataset. To this end, we introduce Eir-8B, an 8-billion-parameter model initially adapted from LLaMA 3.1 Instruct-8B \cite{LLMA3}. We evaluated the model’s grasp of the Thai language using ThaiExam \cite{thaillm-leaderboard}, a benchmark based on Thai language exams. Furthermore, we explored fine-tuning LLaMA 3.1 Instruct-8B to follow Thai-language instructions, enhancing its usability and performance in Thai-specific tasks.

We compared instruction-tuned models using Thai translation datasets and medical terminology, evaluating their Zero-Shot capabilities in tasks such as machine translation, abstractive summarization, and question answering. Our initiative focuses on the development of Eir-8B, a specialized model tailored to the medical domain. This model undergoes extensive pre-training to enhance its medical knowledge and develop a deep understanding of both Thai medical terminology and English. Our ultimate goal is to advance personalized healthcare, or precision medicine, in Thailand by creating a model uniquely suited to the needs of the Thai healthcare context.

\section{Related Work}
In recent years, since the launch of Chat GPT, the potential of natural language processing (NLP) using Transformer-based models has been effectively demonstrated. These models utilize large-scale datasets from various sources such as Wikipedia, BooksCorpus, Common Crawl (CC), and MC4 to generate text that is realistic and natural. Currently, there has been continuous development of models that support the Thai language. However, when considering applications in specific domains like healthcare in Thai, the existing works are still inadequate to meet the needs. Relevant works in this area include:

\subsection{Typhoon: Thai Large Language Models}
Typhoon\cite{Typhoon} is a series of large language models (LLMs) developed specifically for the Thai language. This report discusses the challenges and insights in developing Thai LLMs, including data preparation, model pretraining, instruction-tuning, and evaluation. Despite limited pretraining data, Typhoon uses continual training to transfer knowledge from a strong LLM. The models are evaluated using ThaiExam, a benchmark based on exams for Thai high-school students and professionals. Typhoon, fine-tuned for Thai instructions, outperforms other open-source Thai language models and achieves performance comparable to GPT-3.5 in Thai, with only 7 billion parameters.

\subsection{OpenThaiGPT}
OpenThaiGPT \cite{OpenThaiGPT} is developed from Llama2 \cite{Llama2}, featuring 7 billion parameters. Its tokenizer has been expanded by adding 24,554 additional Thai tokens to enhance text generation performance in the Thai language. The model has been further trained on Thai language data and command datasets that have been translated. While the model's weights have been made publicly available, additional details are limited.

\subsection{Medical Focused LLMs}
\par In the field of medical language models, such as those for Japanese and Chinese, their performance metrics (MedBench) are notable. BioMistral\cite{BioMistral}, released in February 2024, has been continuously trained on medical documents (3 billion tokens) for 1.5 epochs using Mistral-7B, which has been fine-tuned with instructions. Currently, the use of Language Learning Models (LLMs) in healthcare is dominated by private models that are often inaccessible. The most effective models in this space are GPT-4\cite{GPT4} and MedPalm-2\cite{Medpalm2}. In the realm of science and biomedical research, recent studies such as DARE\cite{DARE}, PubMedBERT\cite{PubMedBERT}, SciBERT\cite{SciBERT}, BioNLP\cite{BIONLP}, BioBERT\cite{BioBERT}, ScholarBERT\cite{Sholarbert}, and BioGPT\cite{BioGPT} have demonstrated the efficacy of using well-curated scientific and biomedical datasets for language modeling, both for classification and generation tasks. Although these models show promising results, they are often smaller and narrower in scope compared to large-scale LLMs like GPT-4 and MedPaLM. Despite the progress, specialized LLM applications in the medical domain, such as enhancing clinical assessments and summarizing complex medical communications, face specific challenges. Notably, all of these models still lack robust support for Thai medical language, which remains a significant and urgent gap due to risks of bias, toxicity, inaccurate representation, and misdirection in healthcare.

\section{Methodology}
The development of Eir-8B model has been a complex process, particularly in the creation of a language model tailored specifically for the medical field in Thai. The Thai language, with its unique linguistic structure, presents a limited availability of domain-specific vocabulary, especially within the medical context. This constraint necessitated innovative approaches to effectively train the model.

One key strategy we employed was the incorporation of transliterated English medical terms into the Thai language model. For instance, terms such as "ventilator" or "intubation" were integrated. This approach enabled the model to better comprehend and process medical terms that are widely used but do not have native Thai equivalents.

Additionally, we relied heavily on the generation of synthetic data. This synthetic data proved crucial in addressing gaps where real-world data was either unavailable or insufficient. By blending information from existing medical documents with newly generated content, we were able to construct a more comprehensive dataset, which significantly improved the model’s accuracy and overall performance.

Although the development process was time-intensive and required substantial effort, the outcomes have been highly rewarding. Eir-8B model has demonstrated strong performance and is now prepared for integration into Thailand’s healthcare system, with the objective of contributing to enhanced medical care across the country.

\subsection{Pretraining Dataset}
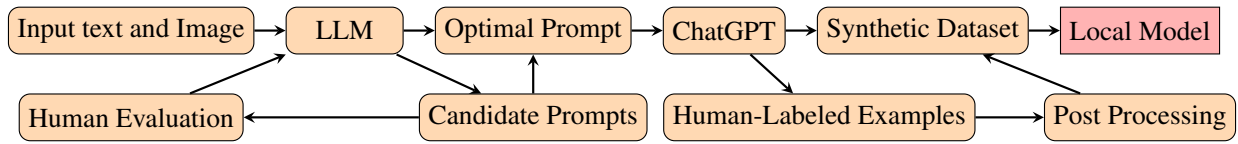
\begin{figure}[h]
\centering
\resizebox{\textwidth}{!}{
\begin{tikzpicture}[node distance=0.1cm and 0.4cm, auto]

  \node (task) [prompt] {Input text and Image};
  \node (llm) [prompt, right=of task] {LLM};
  \node (optprompt) [process, right=of llm] {Optimal Prompt};
  \node (chatgpt) [process, right=of optprompt] {ChatGPT};
  \node (dataset) [process, right=of chatgpt] {Synthetic Dataset};
  \node (localmodel) [data, right=of dataset] {Local Model};
  \node (evaluation) [prompt, below=0.5cm of task] {Human Evaluation};
  \node (candidates) [process, below=0.5cm of optprompt] {Candidate Prompts};
  \node (examples) [process, right = 0.2cm of candidates] {Human-Labeled Examples};
  \node (post) [process, below=0.5cm of localmodel] {Post Processing};

  \draw [arrow] (task) -- (llm);
  \draw [arrow] (llm) -- (optprompt);
  \draw [arrow] (optprompt) -- (chatgpt);
  \draw [arrow] (chatgpt) -- (dataset);
  \draw [arrow] (dataset) -- (localmodel);
  \draw [arrow] (llm) -- (candidates);
  \draw [arrow] (candidates) -- (optprompt);
  \draw [arrow] (candidates) -- (evaluation);
  \draw [arrow] (evaluation) -- (llm);
  \draw [arrow] (chatgpt) -- (examples);
  \draw [arrow] (examples) -- (post);
  \draw [arrow] (post) -- (dataset);

\end{tikzpicture}
}
\caption{An overview of the workflow for synthetic data generation using ChatGPT.}
\label{fig:workflow}

\end{figure}

We have decided to enhance the quality of responses in Thai by utilizing a dataset consisting of detailed, high-quality clinical healthcare content. This dataset comprises 100,000 pages, evenly balanced between Thai and English (50/50). Additionally, we incorporated ICD-10 knowledge to improve the model's understanding of Thai medical contexts and terminology. In cases where specific sections of the dataset required augmentation, we applied synthetic data generation techniques to ensure the highest possible accuracy. This process is outlined in Figure~\ref{fig:workflow}.

\subsection{Dataset}

\subsubsection{Medical Question Answering}
Medical Question Answering: This involves reading comprehension skills and serves as a standard in the Open Medical-LLM Leaderboard for assessing the capabilities of LLMs in the medical domain. The datasets used include MedQA\cite{PubMedQA}, MedMCQA\cite{MedMCQA}, PubMedQA\cite{PubMedQA}, MMLU medical-subset\cite{MMLU}, which are related to medicine and biology. These datasets are crucial for research and include assessments of professional medical knowledge, such as medical exam questions and questions that require comprehension of medical research.

\textbf{MedQA} dataset contains multiple-choice questions from the USMLE, assessing general medical knowledge and reasoning skills for US medical licensure. It includes 11,450 questions in the development set and 1,273 in the test set, with each question offering 4 or 5 answer choices.

\begin{itemize}
  \item \textbf{Format:} Q + A, multiple choice, open domain
  \item \textbf{Question:} A 65-year-old man with hypertension comes to the physician for a routine health maintenance examination. Current medications include atenolol, lisinopril, and atorvastatin. His pulse is 86/min, respirations are 18/min, and blood pressure is 145/95 mm Hg. Cardiac examination reveals end diastolic murmur. Which of the following is the most likely cause of this physical examination?
  \item \textbf{Answer:} (A) Decreased compliance of the left ventricle (B) Myxomatous degeneration of the mitral valve (C) Inflammation of the pericardium (D) Dilation of the aortic root (E) Thickening of the mitral valve leaflet
\end{itemize}

\textbf{MedMCQA} is a large QA dataset from Indian medical entrance exams (AIIMS/NEET), covering 2.4k healthcare topics and 21 medical subjects. It includes over 187,000 questions in the development set and 6,100 questions in the test set, each with 4 answer choices and an explanation. It assesses a model's medical knowledge and reasoning. 

\begin{itemize}
  \item \textbf{Format:} Q + A, multiple choice, open domain
  \item \textbf{Question:} Which of the following ultrasound findings has the highest association with aneuploidy?
  \item \textbf{Answer:} (A) Choroid plexus cyst (B) Nuchal translucency (C) Cystic hygroma (D) Single umbilical artery
\end{itemize}
\begin{itemize}
  \item \textbf{Explanation:} All the above mentioned are ultrasound findings associated with increased risk of aneuploidy although the highest association is seen with cystic hygroma. Nuchal translucency and cystic hygroma are both measured in the first trimester. Trisomy 21 is the most common aneuploidy associated with increased NT and cystic hygroma while monosomy X presents as second-trimester hygroma.
\end{itemize}

\textbf{PubMedQA} is a closed-domain QA dataset with 1,000 expert-labeled question-answer pairs, each linked to a PubMed abstract. The task is to provide a yes/no/maybe answer based on the abstract. The dataset is split into 500 questions for development and 500 for testing, assessing a model's comprehension and reasoning over scientific biomedical literature. 

\begin{itemize}
  \item \textbf{Format:} Q + A + context, multiple choice, closed domain 
  \item \textbf{Question:} Double balloon enteroscopy: is it efficacious and safe in a community setting?
  \item \textbf{Context:} From March 2007 to January 2011, 88 DBE procedures were performed on 66 patients. Indications included evaluation anemia/gastrointestinal bleed, small bowel IBD and dilation of strictures. Video-capsule endoscopy (VCE) was used prior to DBE in 43 of the 66 patients prior to DBE evaluation. The mean age was 62 years. Thirty-two patients were female, 15 were African-American; 44 antegrade and 44 retrograde DBEs were performed. The mean time per antegrade DBE was 107.4 ± 30.0 minutes with a distance of 318.4 ± 152.9 cm reached past the pylorus. The mean time per lower DBE was 100.7 ± 27.3 minutes with 168.9 ± 109.1 cm meters past the ileocecal valve reached. Endoscopic therapy in the form of electrocautery to ablate bleeding sources was performed in 20 patients (30.3\%), biopsy in 17 patients (25.8\%) and dilation of Crohn’s-related small bowel strictures in 4 (6.1\%). 43 VCEs with pathology noted were performed prior to DBE, with findings endoscopically confirmed in 32 cases (74.4\%). In 3 cases the DBE showed findings not noted on VCE.
  \item \textbf{Answer:} Yes 
\end{itemize}
\begin{itemize}
  \item \textbf{Long Answer:} DBE appears to be equally safe and effective when performed in the community setting as compared to a tertiary referral center with a comparable yield, efficacy, and complication rate.
\end{itemize}

\textbf{MMLU} benchmark (Measuring Massive Multitask Language Understanding) includes multiple-choice questions across various domains. For the Open Medical-LLM Leaderboard \cite{openlifescienceai} \cite{singhal2022large}, the relevant medical subsets are:
\begin{itemize}
  \item \textbf{Clinical Knowledge}: 265 questions on clinical knowledge and decision-making.
      \begin{itemize}
      \item \textbf{Question:} The following are features of Alzheimer’s disease except:
      \item \textbf{Answer:} 
      \begin{itemize}
        \item (A) short-term memory loss.
        \item (B) confusion.
        \item (C) poor attention.
        \item \textbf{(D) drowsiness}.
      \end{itemize}
    \end{itemize}
  \item \textbf{Medical Genetics}: 100 questions on medical genetics.
    \begin{itemize}
      \item \textbf{Question:} The allele associated with sickle cell anemia apparently reached a high frequency in some human populations due to:
      \item \textbf{Answer:} 
      \begin{itemize}
        \item (A) random mating
        \item \textbf{(B) superior fitness of heterozygotes in areas where malaria was present}
        \item (C) migration of individuals with the allele into other populations
        \item (D) a high mutation rate at that specific gene.
      \end{itemize}
    \end{itemize}
  \item \textbf{Anatomy}: 135 questions on human anatomy.
      \begin{itemize}
      \item \textbf{Question:} Which of the following controls body temperature, sleep, and appetite?
      \item \textbf{Answer:} 
      \begin{itemize}
        \item (A) Adrenal glands
        \item \textbf{(B) Hypothalamus}
        \item (C) Pancreas
        \item (D) Thalamus
      \end{itemize}
    \end{itemize}
  \item \textbf{Professional Medicine}: 272 questions on professional medical knowledge.
      \begin{itemize}
      \item \textbf{Question:} A 19-year-old woman noticed a mass in her left breast 2 weeks ago while doing monthly breast self-examination. Her mother died of metastatic breast cancer at the age of 40 years. Examination shows large dense breasts; a 2-cm, firm, mobile mass is palpated in the upper outer quadrant of the left breast. There are no changes in the skin or nipple, and there is no palpable axillary adenopathy. Which of the following is the most likely diagnosis?
      \item \textbf{Answer:} 
      \begin{itemize}
        \item \textbf{(A) Fibroadenoma}
        \item (B) Fibrocystic changes of the breast
        \item (C) Infiltrating ductal carcinoma
        \item (D) Intraductal papilloma
      \end{itemize}
    \end{itemize}
  \item \textbf{College Biology}: 144 questions on college-level biology.
      \begin{itemize}
      \item \textbf{Question:} Which of the following is the most direct cause of polyteny in somatic cells of certain organisms?
      \item \textbf{Answer:} 
      \begin{itemize}
        \item (A) RNA transcription
        \item (B) Supercoiling of chromatin
        \item \textbf{(C) Chromosome replication without cell division}
        \item (D) Chromosome non-disjunction.
      \end{itemize}
    \end{itemize}
  \item \textbf{College Medicine}: 173 questions on college-level medical knowledge
      \begin{itemize}
      \item \textbf{Question:} The main factors determining success in sport are:
      \item \textbf{Answer:} 
      \begin{itemize}
        \item (A) a high energy diet and large appetite.
        \item (B) high intelligence and motivation to succeed.
        \item (C) a good coach and the motivation to succeed.
        \item \textbf{(D) innate ability and the capacity to respond to the training stimulus}.
      \end{itemize}
    \end{itemize}
\end{itemize}

\subsubsection{Quality Dataset Standard Questions in Thai Language}
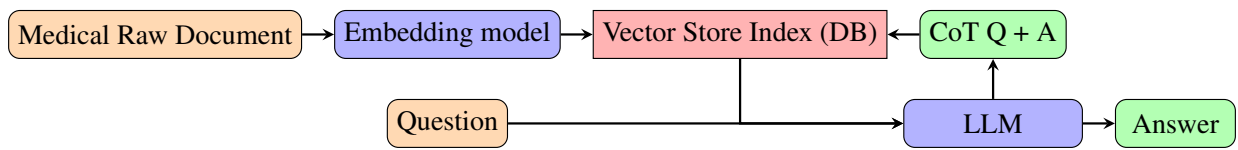
\begin{figure}[ht]
\centering
\resizebox{\textwidth}{!}{
\begin{tikzpicture}[node distance=0.1cm and 0.4cm, auto]

\node (question) [process] {Medical Raw Document};
\node (embedding) [process, right=of question, fill=blue!30] {Embedding model};
\node (question1) [process, below = 0.5 cm of embedding] {Question};
\node (vectorstore) [data, right=of embedding] {Vector Store Index (DB)};
\node (cotqa) [cotqa, right=of vectorstore] {CoT Q + A};
\node (llm) [llm, below= 0.5cm of cotqa, text width=2cm] {LLM};
\node (answer1) [answer, right=of llm] {Answer};

\draw [arrow_cot] (question) -- (embedding);
\draw [arrow_cot] (embedding) -- (vectorstore);
\draw [arrow_cot] (cotqa) -- (vectorstore);
\draw [arrow_cot] (vectorstore) |- (llm);
\draw [arrow_cot] (question1) -- (llm);
\draw [arrow_cot] (llm) -- (cotqa);
\draw [arrow_cot] (llm) -- (answer1);

\end{tikzpicture}
}
\caption{Overview Medical Prompt Workflow for Document Processing and Querying Using Embedding Model}
\label{fig:RagFlow}
\end{figure}

We have developed additional question sets derived from high-quality standard datasets in Thailand using the Retrieval-Augmented Generation (RAG) technique. This approach not only improves the comprehensiveness of question and answer generation but also ensures that the information is accurate and up-to-date. By employing RAG, we are able to integrate data from various sources to create question and answer sets that effectively address user needs. The specifics of this methodology and the resulting data are detailed comprehensively in Figure~\ref{fig:RagFlow}.

We have augmented our medical synthetic data for instruction tuning by creating an additional 266,080 question-answer pairs. These have been incorporated into both the synthetic dataset and the final supervised training data. This comprehensive dataset includes medical and general fine-tuning data, along with the synthetic data, thereby enhancing the model's medical knowledge and its understanding of the Thai language context.

\subsubsection{Data Filter Pipeline}
\begin{figure}[ht]
\centering
\resizebox{\textwidth}{!}{
  \begin{tikzpicture}
    \begin{axis}[
      width=\textwidth,
      height=0.3\textheight,
      ybar,
      bar width=20pt,
      ymin=0,
      ymax=800,
      enlarge x limits=0.1,
      ylabel={Frequency},
      xlabel={Quality Scoring},
      xtick={0, 1, 2, 3, 4, 5, 6, 7, 8, 9, 10},
      nodes near coords,
      every node near coord/.append style={font=\footnotesize, anchor=south},
      title={Frequency Thai Exam Dataset of Quality Scoring Values}
    ]
    \addplot [fill=lime!50] coordinates {(0,5) (1,1) (2,7) (3,17) (4,15) (5,36) (6,33) (7,35) (8,328) (9,187) (10,732)};
    \draw[red, very thick] (axis cs:7.5,0) -- (axis cs:7.5,800);
    \end{axis}
  \end{tikzpicture}
  }
  \caption{Distribution of Thai exam dataset pairs with quality scores, illustrating that the majority received high scores. Data points with scores below 7 (left of the red line) were excluded from training.}
  \label{fig:thai_filter}
\end{figure}
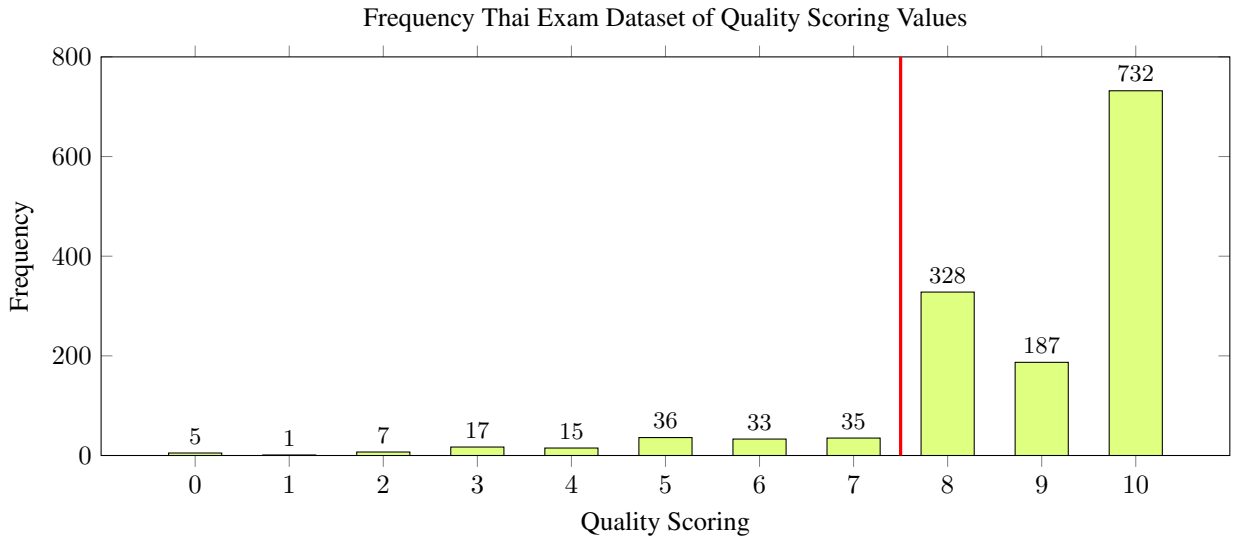

We applied the DEITA \cite{DEITA} technique to evaluate scores on a scale of 0 to 10, as shown in Figures \ref{fig:med_filter} and \ref{fig:thai_filter}. Subsequently, all scores below 7 were removed to filter out low-quality data, ensuring that the analysis and assessments remain accurate and reliable.

\begin{figure}[ht]
\centering
\resizebox{\textwidth}{!}{
  \begin{tikzpicture}
    \begin{axis}[
      width=\textwidth,
      height=0.3\textheight,
      ybar,
      bar width=20pt,
      ymin=0,
      ymax=180000,
      enlarge x limits=0.1,
      ylabel={Frequency},
      xlabel={Quality Scoring},
      xtick={0, 1, 2, 3, 4, 5, 6, 7, 8, 9, 10},
      nodes near coords,
      every node near coord/.append style={font=\footnotesize, anchor=south},
      title={Frequency Medical Dataset of Quality Scoring Values}
    ]
    \addplot [fill=lime!50] coordinates {(0,23725) (1,309) (2,8857) (3,23434) (4,46071) (5,41261) (6,39413) (7,49763) (8,162771) (9,101562) (10,148903)};
    \draw[red, very thick] (axis cs:7.43,0) -- (axis cs:7.43,180000);
    \end{axis}
  \end{tikzpicture}
  }
  \caption{Distribution of medical dataset pairs with quality scores, illustrating that the majority received high scores. Data points with scores below 7 (left of the red line) were excluded from training.}
  \label{fig:med_filter}
\end{figure}
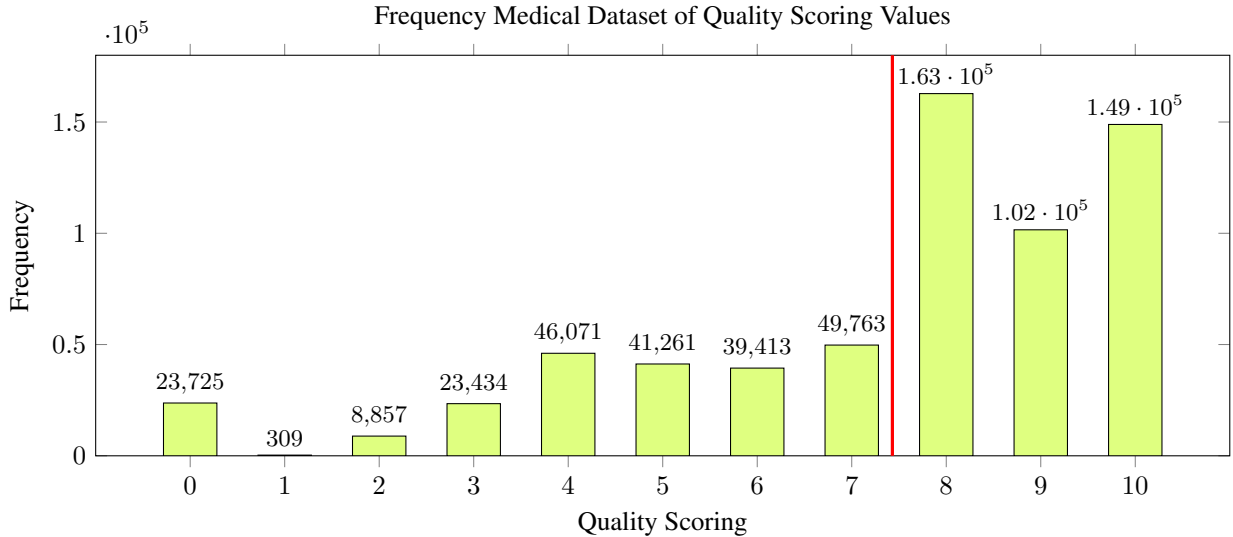

\subsubsection{Thai Medical EHR Data Analysis}
The application of Artificial Intelligence (AI) in the analysis and processing of Electronic Health Records (EHR), alongside its integration into Hospital Information Systems (HIS), aims to improve the accuracy, efficiency, and usability of the medical information stored in these systems. AI models must be trained to recognize patterns and structures within hospital data, including patient records, medical histories, and prescriptions, in order to effectively assist healthcare professionals.

\textbf{Eir-8B}'s role in EHR systems encompasses a variety of tasks, including Named Entity Recognition (NER), which focuses on identifying and categorizing key information such as patient names, medical conditions, medications, and other essential data. Additionally, AI can assist in patient data management, medical analysis, diagnosis, and treatment outcome predictions, enabling healthcare professionals to make faster and more accurate decisions.

To support these objectives, we have identified 18 key areas (Table \ref{tab:ehr_tasks}) where AI can be effectively applied, providing a practical framework for integrating AI into various hospital operations. These categories help tailor AI applications to specific healthcare needs, ensuring seamless and efficient workflows.

In summary, integrating AI with HIS not only reduces the workload of healthcare staff but also enhances the quality of patient care, thereby making healthcare delivery more efficient and precise.

\begin{table}[ht]
  \centering
  \resizebox{\textwidth}{!}{
  \begin{tabular}{|m{3cm}|m{4cm}|m{2cm}|m{5cm}|}
    \hline
    \textbf{Dataset} & \textbf{Format} & \textbf{Size} & \textbf{Domain} \\ \hline
    Neuro & Q + A (Long Answer) & 11,580 & Neurology Clinical Practice \\ \hline
    Kidney & Q + A (Long Answer) & 11,935 & Kidney Exam \\ \hline
    Cardio & Q + A (Long Answer) & 12,150 & Heart Disease \\ \hline
    Chest & Q + A (Long Answer) & 8,200 & Medical knowledge Respiratory Medicine \\ \hline
    Endo & Q + A (Long Answer) & 10,860 & Endocrinology \\ \hline
    Medicine & Q + A (Long Answer) & 70,755 & Medicine knowledge \\ \hline
    Hema & Q + A (Long Answer) & 3,600 & Hematology knowledge \\ \hline
    PsyQA & Q + A (Long Answer) & 17,350 & Psychiatry knowledge \\ \hline
    InfQA & Q + A (Long Answer) & 22,550 & Infectious disease \\ \hline
    MedComQA & Q + A (Long Answer) & 22,550 & Medicine knowledge \\ \hline
    GynQA & Q + A (Long Answer) & 14,300 & Gynecology knowledge \\ \hline
    NurDQA & Q + A (Long Answer) & 7,200 & Nursing Careplan And Intervention \\ \hline
    OncQA & Q + A (Long Answer) & 7,200 & Cancer \\ \hline
    GasQA & Q + A (Long Answer) & 13,450 & Gastroenterology and Hepatology \\ \hline
    SurQA & Q + A (Long Answer) & 11,900 & Surgery \\ \hline
  \end{tabular}
  }
  \vspace{0.5cm}
    \caption{Summary of Thai Medical QA describing the format, size, and domain of the datasets}
    \label{tab:multimedqa_sum}
\end{table}

In addition, we have developed a predictive model and advice system \cite{HealthLLM} enhanced with specialized skills for patient assessment and analysis, tailored to increase specific medical knowledge within Thailand. This model utilizes the Open PMC Patient dataset, focusing on key medical areas such as ICD-10, diagnosis, treatment plans, nursing diagnosis, and discharge summaries, all based on Thai language input. The adaptation process begins with pre-training on clinical notes \cite{SyntheticClinical}, enabling the model to learn the language and patterns of medical documentation. It is then fine-tuned on clinical instructions to improve its ability to understand and generate accurate clinical guidelines and protocols. This comprehensive enhancement aims to improve patient evaluations and healthcare outcomes in the Thai context. The details of this process are outlined in Table \ref{tab:multimedqa_sum} and Table \ref{tab:thai_exam}.

Addressing the challenge of utilizing real patient data in predictive models and health advice for medical professionals and patients requires leveraging publicly available patient datasets. Specifically, the Open PMC Patient and Open-Patient datasets are invaluable due to their extensive size and the wide range of diseases they encompass. Since these datasets are predominantly in English, a critical first step is employing the GPT-4o model to translate this data into Thai. This translation is essential for enabling comprehensive analysis and application of the data in a localized context, thereby enhancing the precision and effectiveness of predictive models and health advice tailored to the Thai medical community.

\clearpage

\begin{table}[ht]
\centering
\begin{tabular}{|p{5cm}|p{10cm}|}
\hline
\textbf{Tasks} & \textbf{Descriptions} \\ \hline
\textbf{Temporal Information Extraction} & Extracts and organizes temporal data from EHRs, such as dates of diagnoses, treatments, or admissions, to track the timeline of patient care. \\ \hline
\textbf{Paraphrasing} & Assesses the ability to rephrase medical information in EHRs while maintaining the original meaning, useful for generating patient summaries or reports. \\ \hline
\textbf{Natural Language Generation (NLG)} & Generates coherent and contextually appropriate text from structured EHR data, such as patient summaries, discharge notes, or medical reports. \\ \hline
\textbf{Keyword Extraction} & Extracts essential keywords from EHRs, facilitating quick access to critical information and improving efficiency in data retrieval. \\ \hline
\textbf{Text Classification} & Categorizes or classifies sections of EHRs into predefined categories, such as diagnosis codes, treatment types, or patient demographics. \\ \hline
\textbf{Relation Extraction} & Identifies and extracts relationships between entities within EHRs, such as the relationship between medications and diagnoses, crucial for understanding patient treatment paths. \\ \hline
\textbf{Question Answering} & Involves answering questions derived from EHR data, such as inquiries about a patient’s medical history or treatment plan, enabling efficient data access for healthcare providers. \\ \hline
\textbf{Text Summarization} & Condenses lengthy EHR data into concise summaries, useful for providing quick overviews of patient histories or treatment outcomes. \\ \hline
\textbf{Abbreviation Expansion} & Expands abbreviations found in EHRs into their full forms, ensuring clarity and reducing ambiguity in medical documentation. \\ \hline
\textbf{Clinical Concept Normalization} & Standardizes medical terminology within EHRs, ensuring consistency in the representation of clinical concepts for accurate data analysis and interoperability. \\ \hline
\textbf{Open-ended Question} & Assesses the ability to generate appropriate responses to open-ended questions based on EHR data, helpful for patient care planning or decision support. \\ \hline
\textbf{Multiple-Choice Question} & Involves selecting the correct answer from multiple choices based on information from EHRs, useful in educational settings or automated decision support systems. \\ \hline
\textbf{Coreference Resolution} & Identifies and resolves references to the same entity within EHRs, ensuring that all mentions of a patient, condition, or treatment are accurately linked. \\ \hline
\textbf{Yes/No Question} & Involves answering binary yes/no questions based on EHR data, aiding in quick decision-making processes. \\ \hline
\textbf{Medical Translate} & Focuses on translating medical information in EHRs from English to Thai, facilitating better communication in multilingual healthcare settings. \\ \hline
\textbf{Medical Thai Extraction} & Extracts medical information specifically from Thai-language EHRs, ensuring that AI systems can process and understand medical data in this language. \\ \hline
\textbf{Medical ICD Prediction} & Predicts International Classification of Diseases (ICD) codes from EHR data, assisting in accurate coding for billing, research, and treatment planning. \\ \hline
\end{tabular}

\vspace{0.5cm}
\caption{Overview of EHR Processing Tasks and Their Descriptions}
\label{tab:ehr_tasks}
\end{table}

\clearpage

\begin{table}[ht]
  \centering
  \resizebox{\textwidth}{!}{
  \begin{tabular}{|m{8cm}|m{5cm}|m{2cm}|}
    \hline
    \textbf{Datasets} & \textbf{Format} & \textbf{Size} \\ \hline
    ALevel Exam ( Chem, Math, Social, Thai ) & Q + A (Choice + Long Answer) & 500 \\ \hline
    IC Exam & Q + A (Choice + Long Answer) & 45 \\ \hline
    TGAT (Math, Social, Thai) & Q + A (Choice + Long Answer) & 119 \\ \hline
    TPAT (Math) & Q + A (Choice + Long Answer) & 223 \\ \hline
    Onet ( Science, Math, Social, Thai ) & Q + A (Choice + Long Answer) & 810 \\ \hline
  \end{tabular}
  }
  \vspace{0.5cm}
    \caption{Summary of Thai Exam QA describing the format, size}
    \label{tab:thai_exam}
\end{table}

\begin{figure}[H]
\centering
\tikzset{
 block/.style = {rectangle, draw, rounded corners, text width=8em, align=center, minimum height=3em},
 arrow/.style = {thick,->,>=stealth},
 dashedarrow/.style = {thick,dashed,->,>=stealth}
}
\begin{tikzpicture}[node distance=0.5cm and 0.8cm, auto]

\node (main) [block, fill=green!30] {MultiAssetment};
\node (consumer) [block, above=of main, fill=green!10] {Consumer Health Analysis};
\node (liveqa) [block, left=of main,draw=black, fill=orange!30] {Nursing Diagnosis};
\node (medqa) [block, right=of main, draw=black, fill=orange!30] {ICD 10};
\node (medication) [block, below left=of main, draw=black, fill=orange!30] {Treatment Plans};
\node (pubmed) [block, below=of main, draw=black, fill=orange!30] { Summary Discharge};
\node (mmlu) [block, below right=of main, draw=black, fill=orange!30] {Health Care Patient};

\draw [arrow] (main) -- (consumer);
\draw [dashedarrow] (liveqa) -- (main);
\draw [dashedarrow] (medqa) -- (main);
\draw [dashedarrow] (medication) -- (main);
\draw [dashedarrow] (pubmed) -- (main);
\draw [dashedarrow] (mmlu) -- (main);

\end{tikzpicture}
\caption{Medical Question Domain Specifics}
\label{fig:ThaiMedicalQuestion}
\end{figure}
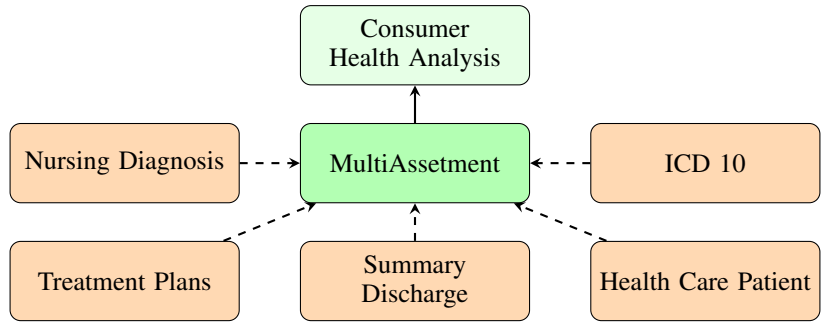

\subsection{Model, Prompt, and Fine-tuning}
This section provides an overview of large language models (LLMs) and the techniques employed to align them with the specific requirements of the medical domain. We begin by detailing the development of the training corpus used during the pre-training phase. Next, we outline the model adaptation methods utilized to refine the model's performance. Finally, we discuss the approaches applied for model merging.

\subsubsection{Model}
The choice of LLaMA 3.1 Instruct-8B\cite{LLMA3} for training models in research is grounded in its superior performance across multiple categories compared to Mistral-7B \cite{Mistral7B} and Gemma 2 9B \cite{Gemma2}. LLaMA 3.1 excels in several key areas: General (MMLU 69.4, outperforming Mistral-7B at 61.1 and Gemma 2 9B at 72.3), Code (HumanEval 72.6, surpassing Mistral-7B at 50.0 and Gemma 2 9B at 54.3), Math (GSM8K 75.6, higher than Mistral-7B at 76.7 and Gemma 2 9B at 84.7), Reasoning (ARC Challenge 83.4, exceeding Mistral-7B at 74.2 and Gemma 2 9B at 87.6), Tool use (BFCL 76.1, better than Mistral-7B at 60.4 and Gemma 2 9B at 70.8), and Multilingual (MGSM 68.9, outperforming Mistral-7B at 29.9 and Gemma 2 9B at 53.2). This comparison demonstrates LLaMA 3.1 Instruct-8B's high efficiency and reliable results, making it an optimal choice for research that demands robust performance and accuracy in handling diverse and complex data.

\subsubsection{Prompt Engineering}
Empirical studies demonstrate that the performance of foundation models in specific tasks can be significantly influenced by prompt engineering. To enhance outcomes, we employed three techniques in combination. Few-shot prompting, in particular, has a substantial impact on model performance. When evaluating GPT-4's efficacy in addressing medical challenges, we constrained the prompts to fundamental contextual learning methods, such as \textbf{one-shot and five-shot prompting} \cite{Cot}, demonstrating how effectively GPT-4 can excel with minimal input.

\textbf{Chain of Thought (CoT)} \cite{Cot} prompting, which uses step-by-step reasoning in sample answers by breaking complex problems into smaller steps, enhances foundational model accuracy. Integrating CoT reasoning steps into few-shot ICL prompts, as seen in Med-PaLM where medical experts crafted CoT prompts for complex medical challenges, improves outcomes. We explored automatic CoT prompt generation using GPT-4 with paired question-answer training data, finding that GPT-4 can autonomously produce high-quality CoT prompts for even the most complex medical problems.

\textbf{Ensembling} \cite{Cot}, which combines results from multiple model runs to produce more accurate or stable outcomes, is achieved through methods such as averaging, voting, or majority voting. This approach further refines performance. The self-consistency ensembling technique employs stochastic methods to generate multiple outputs, which are then aggregated to form a majority decision. Adjusting the "temperature" parameter allows for control over output diversity, with higher temperatures introducing more randomness. By reordering or reconfiguring few-shot prompt components, ensembling mitigates order sensitivity in foundation models, thereby enhancing output stability.

These combined techniques have significantly improved performance on medical benchmarks, including MedQA, MedMCQA, PubMedQA, and the MMLU medical subset, leading to superior outcomes. The specific prompts used can be found in Appendix~\ref{appendix:A}.

\subsubsection{Training Details}

In this subsection, we outline the training process used to fine-tune the pretrained model, LLaMA 3.1 Instruct-8B, utilizing the Low-Rank Adaptation (LoRa) technique \cite{hu2021loralowrankadaptationlarge}. The training details cover the model architecture, dataset preparation, training configuration, and evaluation criteria.

\paragraph{Model Architecture:}The base model used in this study is LLaMA 3.1 Instruct-8B, a large language model with 8 billion parameters. The model's architecture is composed of 32 transformer layers, each with self-attention mechanisms and feed-forward neural networks \cite{LLMA3}. We employed LoRa to reduce the number of trainable parameters by factorizing the weight matrices into low-rank representations. Specifically, the rank $r$ of the LoRa matrices was set to 256.

\paragraph{Dataset Preparation:}For fine-tuning, we utilized a dataset composed of various Thai and English medical content and Thai exams. The dataset was preprocessed by removing low-scoring entries, as mentioned previously, and shuffling the data. The training set consisted of 266,080 samples, and the text was tokenized using a vocabulary size of 2,048 tokens.

\paragraph{Training Configuration:}The model was trained using the following configuration:
\begin{itemize}
  \item \textbf{Optimizer:} We used the \texttt{AdamW} optimizer from \texttt{torch} with a learning rate of $2 \times 10^{-5}$.
  \item \textbf{Batch Size:} The batch size per device was set to 1, with gradient accumulation set to 8 and gradient checkpointing enabled. As there were 4 GPUs, the global batch size was 32.
  \item \textbf{Epochs:} The model was trained for 4 epochs. Thus, total sample is 907,232 samples.
  \item \textbf{Learning Rate Scheduler:} A linear decay learning rate scheduler was employed, with a warm-up phase comprising 10\% of the total training steps.
  \item \textbf{LoRa Hyperparameters:}
  \begin{itemize}
    \item \textbf{Rank ($r$):} The rank of the LoRa matrices was set to 256.
    \item \textbf{Alpha ($\alpha$):} The scaling factor $\alpha$ for the LoRa adaptation was set to 256.
    \item \textbf{Target Modules:} The LoRa adaptation was applied to all linear layers (\texttt{`all-linear'}) in the transformer architecture.
    \item \textbf{Rank Stabilization:} To maintain stability during training, rank stabilization was employed, ensuring that the rank of the LoRa matrices did not degrade over time \cite{kalajdzievski2023rankstabilizationscalingfactor}.
  \end{itemize}
  \item \textbf{DeepSpeed Configuration:} To efficiently train the model across multiple GPUs, we utilized DeepSpeed Zero \cite{rajbhandari2020zeromemoryoptimizationstraining} to manage and optimize GPU resources during the SFT (Supervised Fine-Tuning) process using Hugging Face’s framework. Specifically, DeepSpeed Zero Stage 2 was employed, which distributes optimizer states and gradients across the available GPUs. Given the model size, Stage 2 was deemed appropriate, as Stage 3—which partitions model parameters—was unnecessary for this workload. All Stage 2 configurations were set to \texttt{`auto'} to optimize performance without manual tuning.
  \item \textbf{Hardware:} Training was conducted on 4 NVIDIA A100 GPUs with 40 GB memory each.
\end{itemize}
\paragraph{Model Merging}
After fine-tuning, we employed a model merging strategy to combine the LLaMA-3.1-8B-instruct model with the fine-tuned model. We used the Spherical Linear Interpolation (SLERP) method to merge the models, allocating 50\% to each, using the \texttt{Mergekit} tool \cite{goddard2024arcee}.
\begin{figure}[ht]
    \centering
    \includegraphics[width=0.9\textwidth]{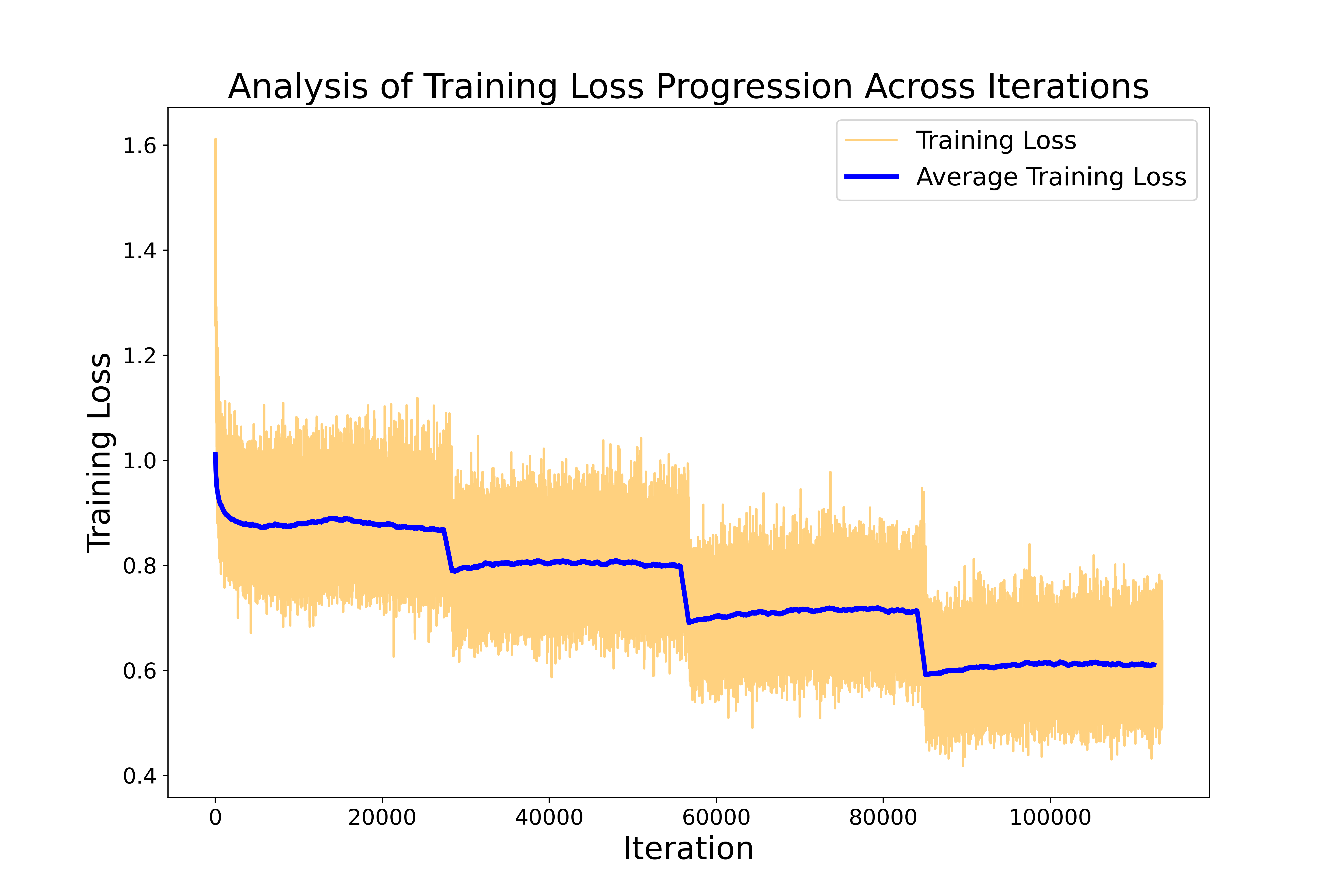}
    \caption{Following four epochs of training, totaling approximately 105 hours, a notable reduction in loss was observed across the epochs, facilitated by the use of rsLoRa. The average loss in the final epoch was 0.608.}
    \label{fig:training_loss}
\end{figure}

\section{Evaluation}
To evaluate the potential of the Eir model, we developed two variants: Eir-8B model, specifically trained for reading Electronic Health Records (EHR) data, and Eir-8B-prob model, which focuses on question-answering tasks. We compared these models with leading medical language models, including commercial LLMs used in Thailand, such as PubMedQA, MedMCQA, MMLU, and MedQA by using Language Model Evaluation Harness \cite{eval-harness}. The test results (Table \ref{tab:table1}) demonstrate that both Eir models outperform the other models in scoring. Notably, Eir-8B-prob model scored 10\% higher than Typhoon-v1.5x-8B-instruct and 14\% higher than GPT-3.5 Turbo 1106, while Eir-8B model achieved a 2.8\% improvement over Typhoon-v1.5x-8B-instruct and a 5.3\% higher score than GPT-3.5 Turbo 1106.


\begin{table}[ht]
\centering
\begin{adjustbox}{max width=\textwidth}
\begin{tabular}{lccccccccccc}
\toprule
\textbf{} & \multicolumn{6}{c}{\textbf{MMLU}} & \textbf{} \\
\cmidrule(lr){2-7}
\textbf{Medical Model} & \textbf{Clinical KG} & \textbf{Medical Genetics} & \textbf{Anatomy} & \textbf{Pro Medicine} & \textbf{College Biology} & \textbf{College Medicine} & \textbf{MedQA} & \textbf{PubMedQA} & \textbf{MedMCQA} & \textbf{Avg.} \\
\midrule
BioMistral 7B \cite{biomistral7B} & 60.9 $\pm$1.5 & 61.7 $\pm$2.1 & 49.6 $\pm$1.2 & 55.1 $\pm$1.3 & 56.9 $\pm$1.5 & 55.5 $\pm$1.7 & 44.4 $\pm$1.2 & 37.6 $\pm$1.5 & 43.9 $\pm$1.0 & 51.7 \\
Mistral 7B Instruct\cite{biomistral7B} & 57.0 $\pm$0.5 & 56.7 $\pm$0.3 & 46.9 $\pm$1.3 & 51.0 $\pm$1.1 & 58.6 $\pm$1.4 & 50.1 $\pm$1.0 & 42.3 $\pm$0.3 & 72.2 $\pm$0.6 & 45.5 $\pm$1.0 & 53.4 \\
MedAlpaca 7B\cite{MedAlpaca} & 49.3 $\pm$0.6 & 48.4 $\pm$1.3 & 39.6 $\pm$1.3 & 48.1 $\pm$0.9 & 63.8 $\pm$0.3 & 45.4 $\pm$1.3 & 35.4 $\pm$0.4 & 56.0 $\pm$0.9 & 39.5 $\pm$0.5 & 47.3 \\
PMC-LLAMA 7B\cite{PMCLLma} & 25.3 $\pm$1.3 & 37.3 $\pm$1.4 & 31.9 $\pm$0.5 & 16.9 $\pm$0.4 & 24.9 $\pm$1.4 & 24.9 $\pm$0.8 & 27.6 $\pm$0.4 & 53.3 $\pm$0.6 & 30.6 $\pm$0.3 & 30.3 \\
MediTron-7B\cite{Meditron} & 37.9 $\pm$1.4 & 47.0 $\pm$0.7 & 44.3 $\pm$0.3 & 31.9 $\pm$1.6 & 36.8 $\pm$1.0 & 45.0 $\pm$2.5 & 34.8 $\pm$1.4 & 59.9 $\pm$0.5 & 41.3 $\pm$0.2 & 42.1 \\
BioMedGPT-LLM-7B \cite{BiomedGPT}& 50.1 $\pm$1.0 & 52.0 $\pm$0.3 & 46.2 $\pm$1.2 & 47.3 $\pm$1.7 & 45.5 $\pm$0.9 & 45.4 $\pm$1.5 & 39.4 $\pm$0.3 & 58.6 $\pm$0.2 & 44.7 $\pm$0.5 & 47.7 \\
GPT-3.5 Turbo 1106 & 74.7 $\pm$0.3 & 60.2 $\pm$2.2 & 65.9 $\pm$2.2 & 72.0 $\pm$1.7 & 64.73 $\pm$2.9 & 64.73 $\pm$2.9 & 57.71 $\pm$1.0 & 72.66 $\pm$1.3 & 66.0 & 66.6 \\
\midrule
\textbf{Thai LLMs} & & & & & & & \\
Eir-8B & 75.1 $\pm$2.7 & 80.0 $\pm$4.0 & 69.6 $\pm$4.0 & 76.8 $\pm$2.6 & 77.1 $\pm$3.5 & 66.5 $\pm$3.6 & 64.5 $\pm$1.3 & \textbf{79.0} $\pm$1.8 & 58.6 $\pm$0.8 & 71.9 \\
Eir-8B + Prob& \textbf{83.8} $\pm$2.3 & \textbf{89.0} $\pm$3.1 & \textbf{83.0} $\pm$3.2 & \textbf{84.9} $\pm$2.2 & \textbf{89.6} $\pm$2.6 & \textbf{75.7} $\pm$3.3 & \textbf{69.6} $\pm$1.3 & 78.8 $\pm$1.8 & \textbf{67.1} $\pm$0.7 & \textbf{80.2} \\

Typhoon-v1.5x-8B-instruct & 75.9 $\pm$2.6 & 79.0 $\pm$4.0 & 63.7 $\pm$4.2 & 70.6 $\pm$2.8 & 77.1 $\pm$3.5 & 63.6 $\pm$3.7 & 59.7 $\pm$1.4 & 74.4 $\pm$2.0 & 58.0 $\pm$0.8 & 69.1 \\
OpenThaiGPT-beta-7B & 37.4 $\pm$3.0 & 38.0 $\pm$4.9 & 4.5 $\pm$4.3 & 32.7 $\pm$2.9 & 36.1 $\pm$4.0 & 32.4 $\pm$3.6 & 32.4 $\pm$1.3 & 62.0 $\pm$2.2 & 31.8 $\pm$0.7 & 34.1 \\
\bottomrule
\end{tabular}
\end{adjustbox}
\vspace{0.5cm}
\caption{Comparing the performance of various medical models, including both commercial and Thai-specific large language models (LLMs), across multiple medical domains. The scores, evaluated using the MMLU benchmark, reflect each model's ability to answer medical questions in fields.}
\label{tab:table1}

\end{table}

To evaluate the ability of the Eir model in handling Thai language, we conducted tests on Thai language data using standard evaluation methods for large language models (LLMs) based on SEACrowd \cite{SeaCrowd}. The primary objective was to assess the model's capability to answer general questions in Thai and verify that it has not undergone significant forgetting or catastrophic knowledge loss. The results (Table \ref{tab:table2}) show that the M3EXAM score reached 0.458, a positive outcome that reflects the model's satisfactory performance in Thai language evaluation.

\begin{table}[ht]
\centering

\begin{adjustbox}{max width=\textwidth}
\begin{tabular}{lcccccccccc}
\toprule
\textbf{Model} &\textbf{ThaiExam} & \textbf{M3Exam} & \textbf{XNLI} & \textbf{XCOPA} \\
\midrule
Eir-8B  & 0.418 & \textbf{0.458} & 0.332 & 0.734 \\
Typhoon-v1.5x-8B-instruct & 0.421 & 0.441 & 0.334 & \textbf{0.822} \\
OpenThaiGPT-beta-7B & 0.253 & 0.284 & \textbf{0.343} & 0.524 \\
Meta Llama 3.1-8B Instruct & \textbf{0.441} & 0.446 & 0.335 & 0.732 \\
\midrule
GPT-3.5-turbo-0613 &  0.460 & 0.341 & 0447 & 0.630 \\
GPT-4-0613 & 0.602 & 0.560 & 0.623 & 0.920 \\
\bottomrule
\end{tabular}
\end{adjustbox}
\vspace{0.5cm}
\caption{Presenting the results of Eir-8B and other models on Thai language evaluation tasks \cite{thaillm-leaderboard}, including ThaiExam, M3Exam, XNLI, and XCOPA. Eir-8B model achieved competitive results, particularly on the M3Exam with a score of 0.458.}
\label{tab:table2}

\end{table}

Additionally, we have created a dataset specifically designed for evaluating medical translation performance. Given that medical terminology in Thailand often includes transliterated terms and may require results in English rather than direct Thai translations, medical professionals may find it challenging to use model outputs effectively. To address this issue, we employed the BLEU score technique for evaluation \cite{BLEU}. As shown in Table \ref{tab:bleu_scores}, Eir-8B model achieved a BLEU score of 61.10 out of 100, the highest among all models tested, with a translation ratio close to 1.0, indicating near-optimal text length.

\begin{table}[ht]
\centering
\begin{tabular}{lcccc}
\toprule
\textbf{Model} & \textbf{BLEU Score} & \textbf{N-gram Precisions (\%)} & \textbf{BP} & \textbf{Ratio} \\
\midrule
Typhoon-v1.5x-8B-Instruct & 34.42 &  71.3/50.6/38.6/29.6 &  0.764 & 0.788 \\
Meta Llama 3.1-8B Instruct & 35.74 &  62.8/42.3/31.7/24.1 &  0.946 & 0.948 \\
Eir-8B & \textbf{61.10} & \textbf{76.1/64.6/56.6/50.1} & \textbf{1.000}  &  \textbf{1.006} \\
Eir-8B-prob & 47.91 & 74.0/58.0/48.2/40.6 & 0.890 &  0.896\\
\bottomrule
\end{tabular}
\vspace{0.5cm}
\caption{Presenting a comparative analysis of various models' performance in medical translation, evaluated using the BLEU score technique. Eir-8B model stands out with a BLEU score of 61.10, the highest among all models tested, showcasing superior translation quality.}
\label{tab:bleu_scores}
\end{table}

Finally, we developed a Clinically Adapted Model Enhanced test using Thai language data to assess how effectively models can be applied in the medical field. This test also provides opportunities for additional training in other areas. The test comprises 18 categories, as shown in Table \ref{tab:table4}. We used GPT-4o to score the answers from four different models on a scale of 0 to 10. Eir-8B model, designed as a Clinically Adapted Model Enhanced, achieved an average score of 7.11, which is 11\% higher than GPT-7o. In 15 of the categories, EIR-8B achieved the highest average score. It also ranked second, just behind GPT-4o, in the categories of Medical Translation, Medical Thai Extraction, and Medical ICD Prediction.

\clearpage
\begin{table}[ht]
\centering
\begin{adjustbox}{max width=\textwidth}
\begin{tabular}{lcccccccccc}
\toprule
\textbf{Task (Number of Q/A)} & \textbf{GPT-3.5} & \textbf{Typhoon-v1.5x-8B-instruct} & \textbf{GPT-4o} & \textbf{Eir-8B } & \\
\midrule
Named Entity Recognition (92) & 3.26 & 5.55 & 6.34 & \textbf{7.08} \\
Temporal Information Extraction (99) & 3.83 & 5.46 & 6.15 & \textbf{7.05} \\
Paraphrasing (7) & 2.36 & 4.68 & 6.35 & \textbf{7.06} \\
Natural Language Generation (86) & 2.63 & 4.87 & 6.91 & \textbf{7.66} \\
Keyword Extraction (75) & 2.60 & 5.15 & 7.01 & \textbf{7.35} \\
Text Classification (76) & 2.92 & 6.21 & 5.36 & \textbf{6.75} \\
Relation Extraction (101) & 3.29 & 5.94 & 4.37 & \textbf{6.92} \\
Question Answering (87) & 3.70 & 4.92 & 6.11 & \textbf{6.82} \\
Text Summarization (97) & 2.98 & 5.44 & \textbf{7.51} & \textbf{7.51} \\
Abbreviation Expansion (67) & 3.99 & 5.96 & 6.24 & \textbf{7.82} \\
Clinical Concept Normalization (78) & 2.67 & 5.63 & 5.82 & \textbf{6.55} \\
Open-ended Question (20) & 3.32 & 5.55 & 6.77 & \textbf{7.27} \\
Multiple-Choice Question (10) & 3.90 & 5.00 & 5.40 & \textbf{6.40} \\
Coreference Resolution (42) & 3.48 & 4.55 & 4.88 & \textbf{6.43} \\
Yes/No Question (21) & 2.71 & 5.86 & 4.86 & \textbf{7.38} \\
Medical Translate (29) & 3.00 & 4.00 & \textbf{7.79} & 6.55 \\
Medical Thai Extraction (63) & 2.81 & 7.16 & \textbf{8.62} & 8.16 \\
Medical ICD Prediction (76) & 2.08 & 3.16 & \textbf{8.12} & 6.41 \\
\midrule
\textbf{Average Score} & 3.05 & 5.33 & 6.38 & \textbf{7.11} \\
\bottomrule
\end{tabular}
\end{adjustbox}
\vspace{0.5cm}
\caption{Providing a comprehensive comparison of the performance of various models—GPT-3.5, Typhoon-v1.5x-8B-instruct, GPT-4o, and Eir-8B across 18 different tasks related to medical language processing. Each task is scored on a scale of 0 to 10.}
\label{tab:table4}
\end{table}

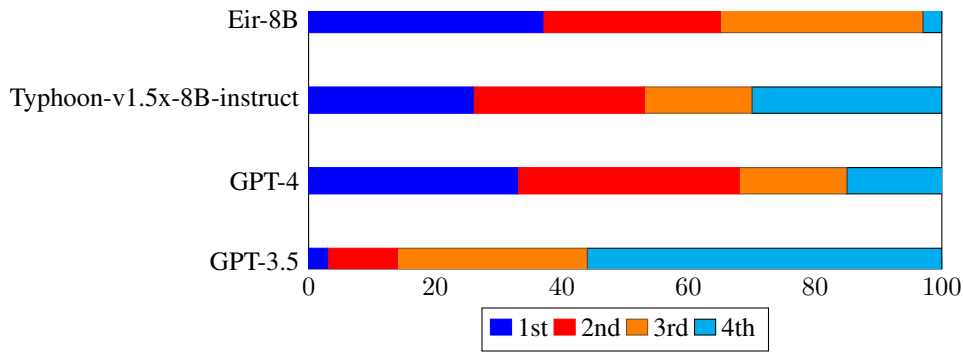
\begin{figure}[ht]
    \begin{tikzpicture}
        \begin{axis}[
            xbar stacked, 
            bar width=0.35cm, 
            width=10cm, 
            height=5cm, 
            xlabel={Percentage (\%)},
            symbolic y coords={GPT-3.5, GPT-4, Typhoon-v1.5x-8B-instruct, Eir-8B},
            ytick=data,
            xmin=0, xmax=100,
            legend style={at={(0.5,-0.15)},anchor=north,legend columns=-1}, 
            enlarge y limits={abs=0.1cm}, 
        ]
        \addplot+[xbar, fill=blue] plot coordinates {(3,GPT-3.5) (33,GPT-4) (26,Typhoon-v1.5x-8B-instruct) (37,Eir-8B)}; 
        \addplot+[xbar, fill=red] plot coordinates {(11,GPT-3.5) (35,GPT-4) (27,Typhoon-v1.5x-8B-instruct) (28,Eir-8B)}; 
        \addplot+[xbar, fill=orange] plot coordinates {(30,GPT-3.5) (17,GPT-4) (17,Typhoon-v1.5x-8B-instruct) (32,Eir-8B)}; 
        \addplot+[xbar, fill=cyan] plot coordinates {(56,GPT-3.5) (16,GPT-4) (30,Typhoon-v1.5x-8B-instruct) (3,Eir-8B)}; 
        \legend{1st, 2nd, 3rd, 4th} 
        \end{axis}
    \end{tikzpicture}
    \caption{Horizontal Stacked Bar Chart of Model Rankings in Normalized Percentages}
    \label{fig:horizontal_stacked_bar_normalized}
\end{figure}
\clearpage
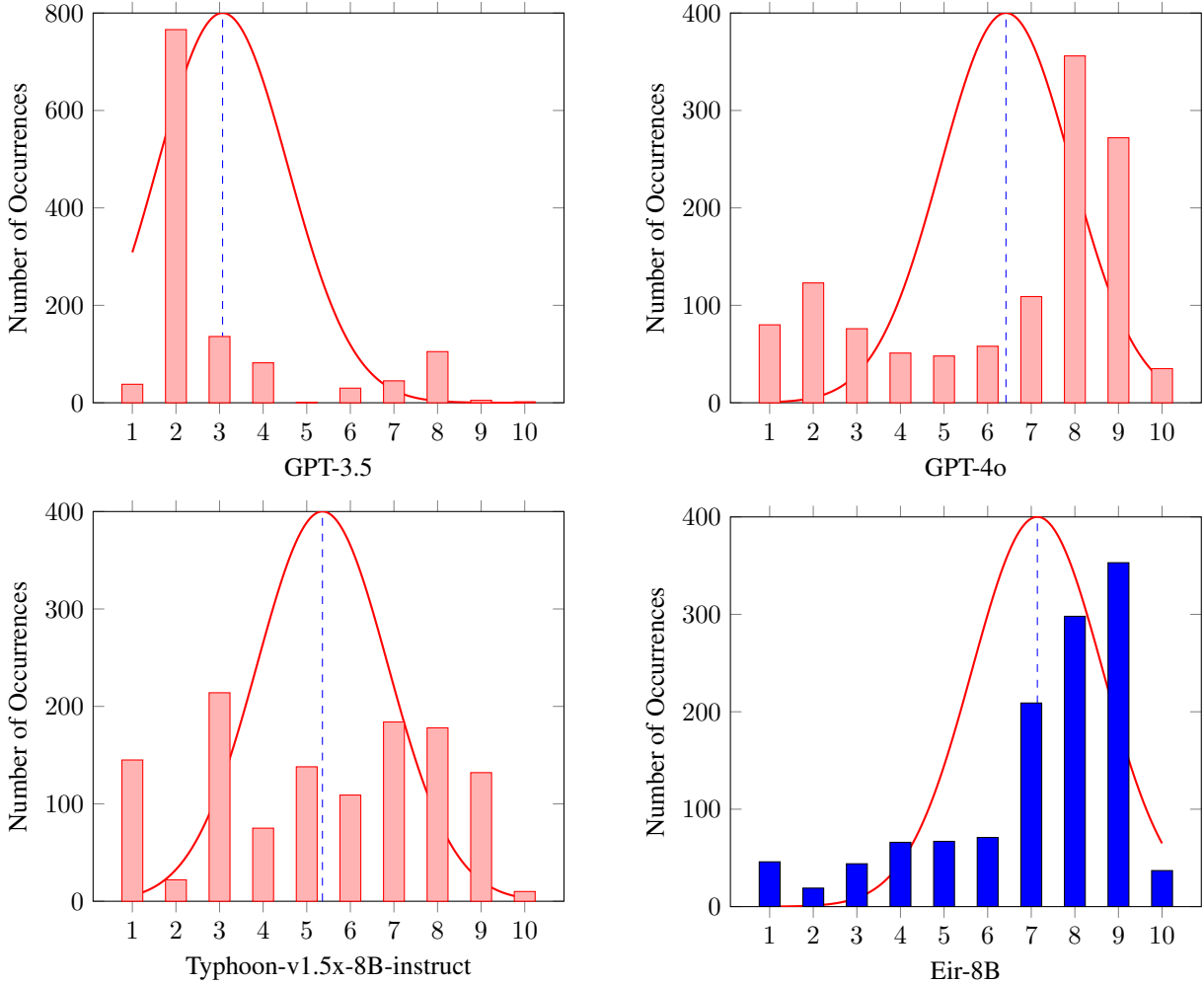
\begin{figure}[ht]
    \centering
    \begin{subfigure}{0.48\textwidth}
        \centering
        \begin{tikzpicture}
        \begin{axis}[
            width=\textwidth,
            ybar,
            xtick={1,2,3,4,5,6,7,8,9,10}, 
            ylabel={Number of Occurrences},
            xlabel={GPT-3.5},
            ymin=0,
            ymax=800,
            bar width=8pt
        ]
        \addplot[red, thick, smooth, no marks, domain=1:10, samples=500] {800 * exp(-((x-3.07)^2)/(2*(1.5)^2))};  
        
        \draw[blue, dashed] (axis cs:3.07,0) -- (axis cs:3.07,800) node[above] {Mean = 3.07};
        
        \addplot coordinates {(1,38) (2,766) (3,136) (4,82) (5,1) (6,30) (7,45) (8,105) (9,5) (10,2)};
        
        \end{axis}
        \end{tikzpicture}
    \end{subfigure}
    \hfill
    \begin{subfigure}{0.48\textwidth}
        \centering
        \begin{tikzpicture}
        \begin{axis}[
            width=\textwidth,
            ybar,
            xtick={1,2,3,4,5,6,7,8,9,10},
            ylabel={Number of Occurrences},
            xlabel={GPT-4o},
            ymin=0,
            ymax=400,
            bar width=8pt
        ]
        \addplot[red, thick, smooth, no marks, domain=1:10, samples=500] {400 * exp(-((x-6.42)^2)/(2*(1.5)^2))};
        
        \draw[blue, dashed] (axis cs:6.42,0) -- (axis cs:6.42,400) node[above] {Mean = 6.42};
        
        \addplot coordinates {(1,80) (2,123) (3,76) (4,51) (5,48) (6,58) (7,109) (8,356) (9,272) (10,35)};
        
        \end{axis}
        \end{tikzpicture}
    \end{subfigure}

    \vspace{0.1cm}

    \begin{subfigure}{0.48\textwidth}
        \centering
        \begin{tikzpicture}
        \begin{axis}[
            width=\textwidth,
            ybar,
            xtick={1,2,3,4,5,6,7,8,9,10},
            ylabel={Number of Occurrences},
            xlabel={Typhoon-v1.5x-8B-instruct},
            ymin=0,
            ymax=400,
            bar width=8pt
        ]
        \addplot[red, thick, smooth, no marks, domain=1:10, samples=500] {400 * exp(-((x-5.36)^2)/(2*(1.5)^2))};
        
        \draw[blue, dashed] (axis cs:5.36,0) -- (axis cs:5.36,400) node[above] {Mean = 5.36};
        
        \addplot coordinates {(1,145) (2,22) (3,214) (4,75) (5,138) (6,109) (7,184) (8,178) (9,132) (10,10)};
        
        \end{axis}
        \end{tikzpicture}
    \end{subfigure}
    \hfill
    \begin{subfigure}{0.48\textwidth}
        \centering
        \begin{tikzpicture}
        \begin{axis}[
            width=\textwidth,
            ybar,
            xtick={1,2,3,4,5,6,7,8,9,10},
            ylabel={Number of Occurrences},
            xlabel={Eir-8B},
            ymin=0,
            ymax=400,
            bar width=8pt
        ]
        \addplot[red, thick, smooth, no marks, domain=1:10, samples=500] {400 * exp(-((x-7.14)^2)/(2*(1.5)^2))};
        
        \draw[blue, dashed] (axis cs:7.14,0) -- (axis cs:7.14,400) node[above] {Mean = 7.14};

        \addplot [fill =blue] coordinates {(1,46) (2,19) (3,44) (4,66) (5,67) (6,71) (7,209) (8,298) (9,353) (10,37)};

        \end{axis}
        \end{tikzpicture}
    \end{subfigure}
\caption{We have separated the histogram into four graphs, with the number of occurrences on the Y-axis and the score on the X-axis. Each graph compares the scores of different models, along with a red curve indicating the distribution trend and a dashed line showing the average score. Eir-8B scores predominantly fall between 7-9, demonstrating that this model performs higher than the others, with a mean score of 7.}
\end{figure}

\section{Conclusion}
\textbf{Eir-8B} is a domain-specific Thai medical LLM that demonstrates advanced medical reasoning and improved performance on specialized benchmarks. Trained continuously on carefully selected, high-quality medical resources, including updated clinical guidelines, Eir-8B outperforms all state-of-the-art models of similar size in Thai medical language capabilities. Notably, it also surpasses all open-source LLMs in both general and medical tasks on Thai-language medical benchmarks.

The model has been released alongside essential tools for managing training datasets and an open-source distributed training library. This approach ensures accessibility for real-world evaluation while enabling further refinements and guided learning.

\textbf{Safety}: Although Eir-8B is designed to encode high-quality medical knowledge, the current online distributed version is not yet optimized for safe, practical use in real-world medical settings. The team strongly advises against using this model for clinical applications without further rigorous testing, including randomized controlled trials. While the current version is not ready for real-world deployment, it is available for researchers to explore the potential of large language models in medical contexts.

\clearpage

\bibliographystyle{plain}
\bibliography{sample}
\newpage
\appendix
\section{Prompting Template}
\label{appendix:A}

We have developed 5-10 templates for each question in the dataset, including additional designs specifically for generating Chain-of-Thought (CoT) answers across various datasets such as MedQA, MedMCQA, PubmedQA, and standardized tests like A-Level, IC, TGAT, and TPAT. These templates play a crucial role in accurately evaluating the model’s responses to both medical questions and academic assessments across diverse fields. Each template is carefully crafted to encompass different question formats and reasoning scenarios, ensuring a thorough assessment of the model’s ability to process and generate effective outcomes.

\begin{figure}[ht]
    \centering
    \includegraphics[width=1\textwidth]{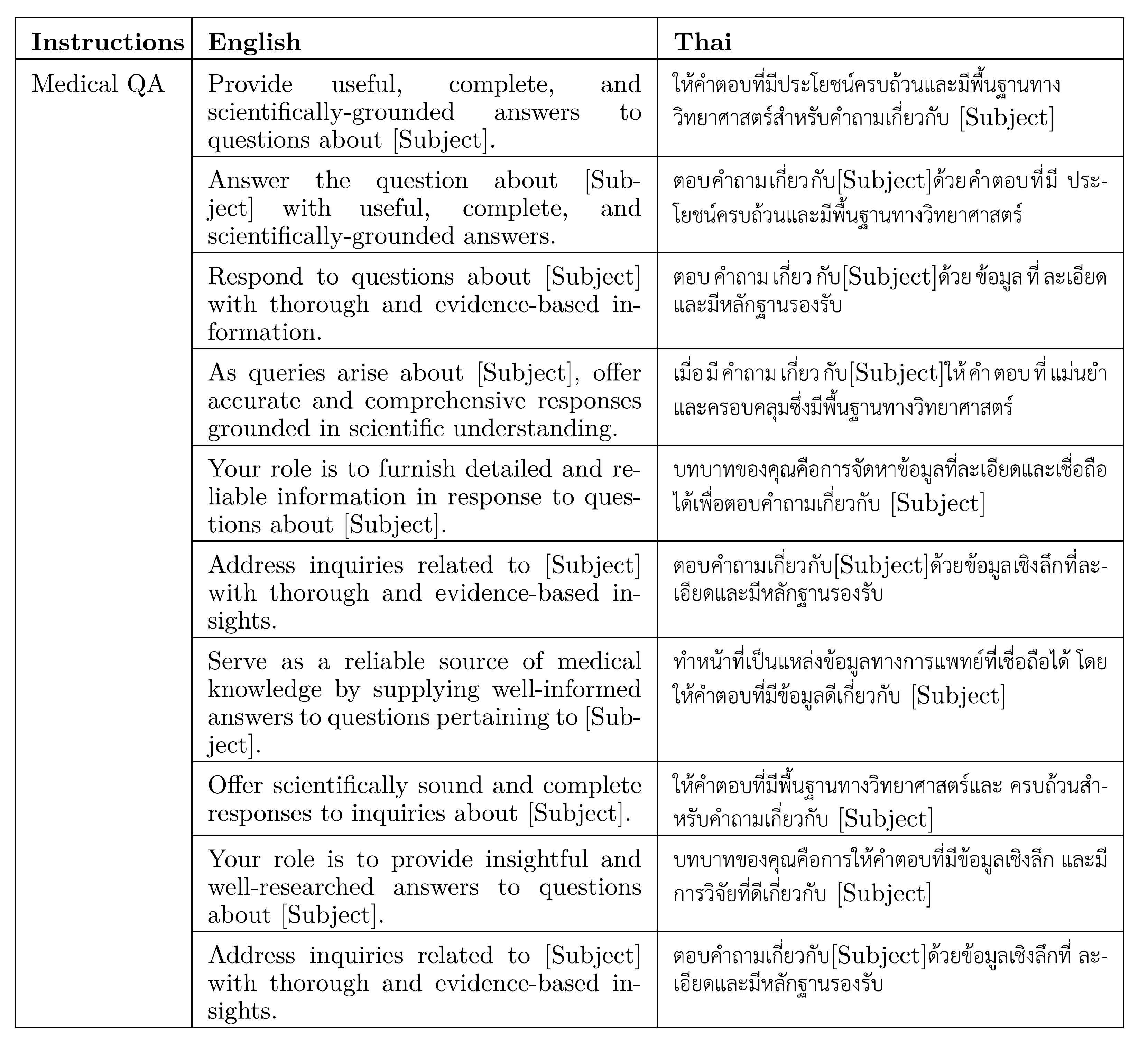}
    \caption{Medical QA}
    \label{fig:Medical QA}
\end{figure}

\begin{figure}[ht]
    \centering
    \includegraphics[width=1\textwidth]{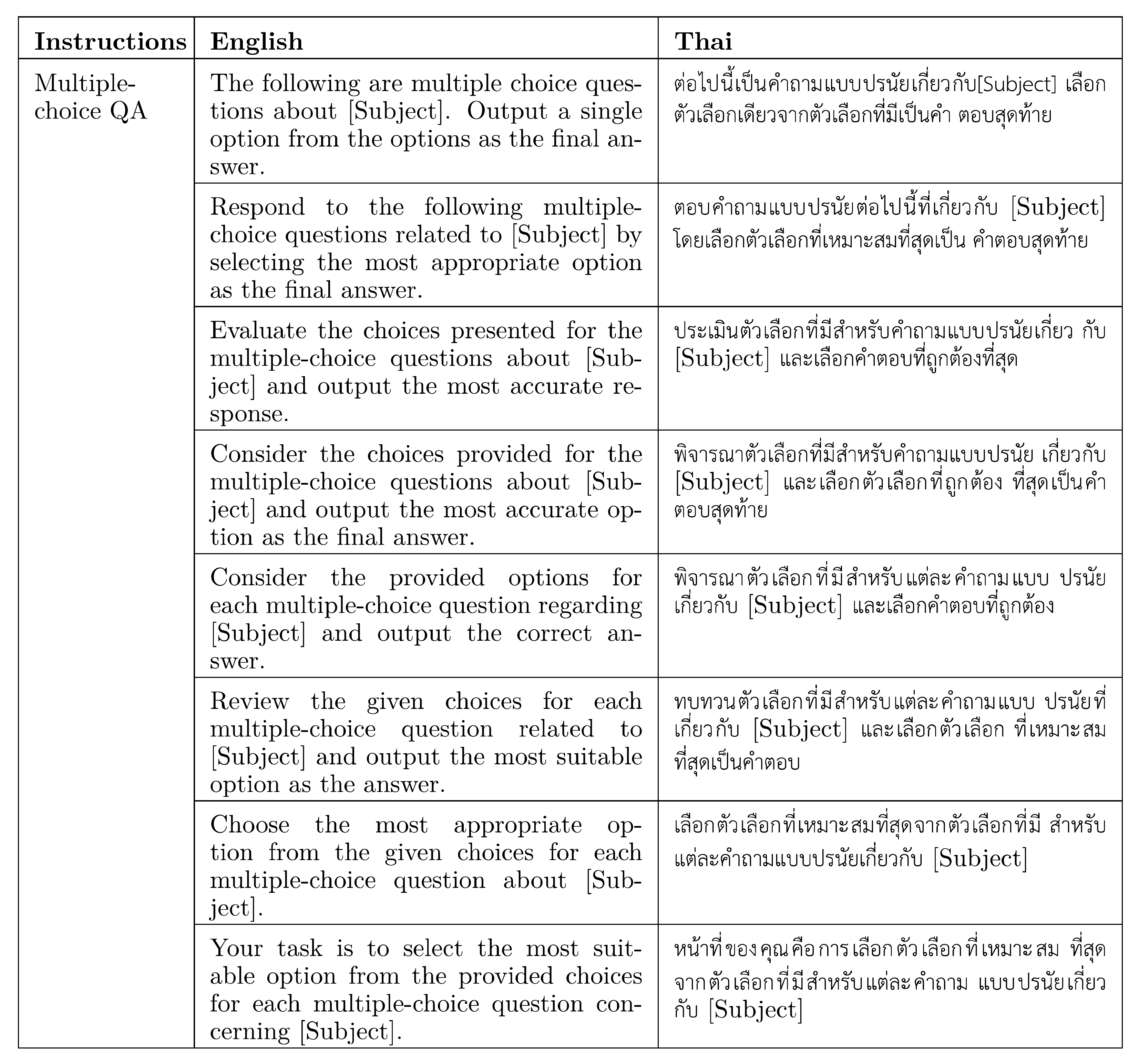}
    \caption{Multiple Choice QA For TPAT , ONET ,IC, ALevel , TGAT.}
    \label{fig:MultipleChoiceThaiExam}
\end{figure}

\clearpage
\begin{table}[ht]
  \centering
  \resizebox{\textwidth}{!}{
  \begin{tabular}{|p{16cm}|}
    \hline
    \textbf{CoT generation (MedQA and MedMCQA)} \\ \hline
    Given the following medical question with options, your task is to select the correct answer by the following process: Summarize the Question: Begin by summarizing the key focus of the question to understand what it’s asking about, such as a specific organ system, pharmacology, or a medical procedure.Analyze Each Option Individually: Carefully review and analyze each option, connecting it to relevant information and medical principles related to the question.Select the Correct Answer: After analyzing all options, determine the best answer by synthesizing the information and reasoning through each choice.Conclude with the Final Answer: Confirm your final answer in the following format: Answer: Option. Explanation.\\ \hline 
  When faced with a medical question with multiple options, the process should begin by first capturing the essence of the question, ensuring a clear understanding of its primary focus. Following this, each option must be independently analyzed, carefully weighing its validity against medical principles and knowledge. After thoroughly evaluating all the options, the correct answer is determined by logically narrowing down the choices, selecting the one that aligns best with the question’s requirements. Finally, the conclusion is stated with clarity, summarizing the selected option alongside a brief explanation to reinforce the reasoning behind the choice. \\ \hline Upon encountering a medical question with multiple options, your approach should involve determining the most accurate response using this process: Begin by giving a concise summary of the scenario to capture its key aspects. Next, thoroughly evaluate each option by considering relevant medical principles and evidence. Ultimately, after systematically analyzing all possibilities, conclude by identifying and selecting the correct answer that best aligns with the context and requirements of the question. \\ \hline
     To answer the medical question effectively, it’s essential to first grasp the core of what is being asked, identifying the primary topic or focus, such as diagnosis, treatment, or pathology. Once the question is clear, each option should be evaluated carefully, connecting it to relevant medical principles and guidelines. This involves analyzing whether each option is plausible within the context provided. As you go through the options, systematically eliminate those that are clearly incorrect or irrelevant, narrowing down the possibilities. The next step is to determine which of the remaining choices best aligns with the scenario based on logical reasoning and clinical evidence. Finally, after selecting the most appropriate answer, it’s important to articulate why this option is correct and how it directly addresses the question at hand. This method ensures a well-rounded and accurate decision-making process. \\ \hline
       Presented with a medical question accompanied by multiple choices, your objective is to identify the correct response employing a systematic strategy. Start by summarizing the essence of the query, then meticulously assess each option in isolation. Conclude by employing a logical and sequential reasoning process to determine the correct answer. Clarify the selected option at the end.\\ \hline
       Encountering a medical inquiry with multiple alternatives, your objective is to determine the correct answer using a systematic methodology. Start with a brief overview of the question’s key focus to establish the context. Next, conduct a detailed analysis of each option, assessing its relevance and accuracy based on established medical knowledge. Finally, apply a step-by-step reasoning process to identify the most accurate answer and clearly state your final selection. \\ \hline
       To approach the given medical question, let’s think through it step by step: First, identify the key elements in the question. Determine what the question is focusing on, such as a specific condition, symptom, treatment, or medical concept. Understanding this will guide us in evaluating the options. Next, analyze each of the given options one by one. For each option, consider whether it logically fits the scenario described in the question. Assess its relevance based on established medical knowledge, clinical guidelines, or common practice. Then, begin eliminating options that are clearly incorrect or don’t align with the core of the question. This helps narrow down the choices, leaving only the most plausible options. After narrowing down, compare the remaining choices closely. Look for subtle differences that make one option more accurate or appropriate than the others. Finally, select the option that best answers the question and provides the most logical and evidence-based solution. Answer: [Selected Option]. This step-by-step approach ensures that the decision is based on careful analysis and sound reasoning. \\ \hline
  \end{tabular}
  }
  \vspace{0.5cm}
  \caption{Cot Promting Generation MedQA and MedMCQA}

\end{table}

\clearpage

\begin{tcolorbox}[title=Medical Generation Prompt]
\textbf{System prompt} : To approach the given medical question, let’s think through it step by step: First, identify the key elements in the question. Determine what the question is focusing on, such as a specific condition, symptom, treatment, or medical concept. Understanding this will guide us in evaluating the options. Next, analyze each of the given options one by one. For each option, consider whether it logically fits the scenario described in the question. Assess its relevance based on established medical knowledge, clinical guidelines, or common practice. Then, begin eliminating options that are clearly incorrect or don’t align with the core of the question. This helps narrow down the choices, leaving only the most plausible options. After narrowing down, compare the remaining choices closely. Look for subtle differences that make one option more accurate or appropriate than the others. Finally, select the option that best answers the question and provides the most logical and evidence-based solution. Answer: [Selected Option]. This step-by-step approach ensures that the decision is based on careful analysis and sound reasoning.\newline
\textbf{Few-shot examples}\newline
For the following multiple-choice question, select one correct answer. Let's think step by step. Question: A 23-year-old pregnant woman at 22 weeks gestation presents with burning upon urination. She states it started 1 day ago and has been worsening despite drinking more water and taking cranberry extract. She otherwise feels well and is followed by a doctor for her pregnancy. Her temperature is 97.7°F (36.5°C), blood pressure is 122/77 mmHg, pulse is 80/min, respirations are 19/min, and oxygen saturation is 98 percent on room air. Physical exam is notable for an absence of costovertebral angle tenderness and a gravid uterus. Which of the following is the best treatment for this patient?\newline
\textbf{Options:} \newline
A. Ampicillin\newline
B. Ceftriaxone \newline
C. Doxycycline\newline
D. Nitrofurantoin\newline
\textbf{Response:} \textbf{Step 1:} Identify Key Elements
The patient is a 23-year-old pregnant woman at 22 weeks gestation with symptoms of burning upon urination. The main focus here is a urinary tract infection (UTI) during pregnancy.
\textbf{Step 2:} Analyze the Options
Each option presents a different antibiotic, and the goal is to determine which is the safest and most effective treatment during pregnancy, considering both efficacy and safety for the fetus.
\begin{enumerate}[label=\Alph*.]
  \item Ampicillin
Ampicillin is a broad-spectrum antibiotic that can be used during pregnancy. However, it is not typically first-line for uncomplicated UTIs.
  \item Ceftriaxone
Ceftriaxone is a third-generation cephalosporin often used for more severe infections like pyelonephritis, not typically for uncomplicated UTIs.
  \item Doxycycline
Doxycycline is contraindicated during pregnancy due to potential teratogenic effects, including fetal bone and teeth abnormalities.
  \item Nitrofurantoin
Nitrofurantoin is a common first-line treatment for uncomplicated UTIs during pregnancy, particularly in the second trimester. It is both effective and safe when used appropriately.
\end{enumerate}

\textbf{Step 3: Eliminate Clearly Incorrect Options}\newline
C. Doxycycline can be eliminated immediately as it is contraindicated during pregnancy.
B. Ceftriaxone is generally reserved for more complicated infections, not a first-line treatment for simple UTIs.
A. Ampicillin is less commonly used as the first-line choice due to resistance issues and is less effective compared to other options like nitrofurantoin.
Step 4: Compare Remaining Choices
Between A. Ampicillin and D. Nitrofurantoin, nitrofurantoin is the preferred option for treating uncomplicated UTIs in pregnant women, especially during the second trimester.
\newline
\textbf{Final Answer
Given the analysis, the best treatment for this patient is:}
\textbf{Answer}: D. Nitrofurantoin
\end{tcolorbox}

\newpage
\begin{figure}[ht]
    \centering
    \includegraphics[width=1\textwidth]{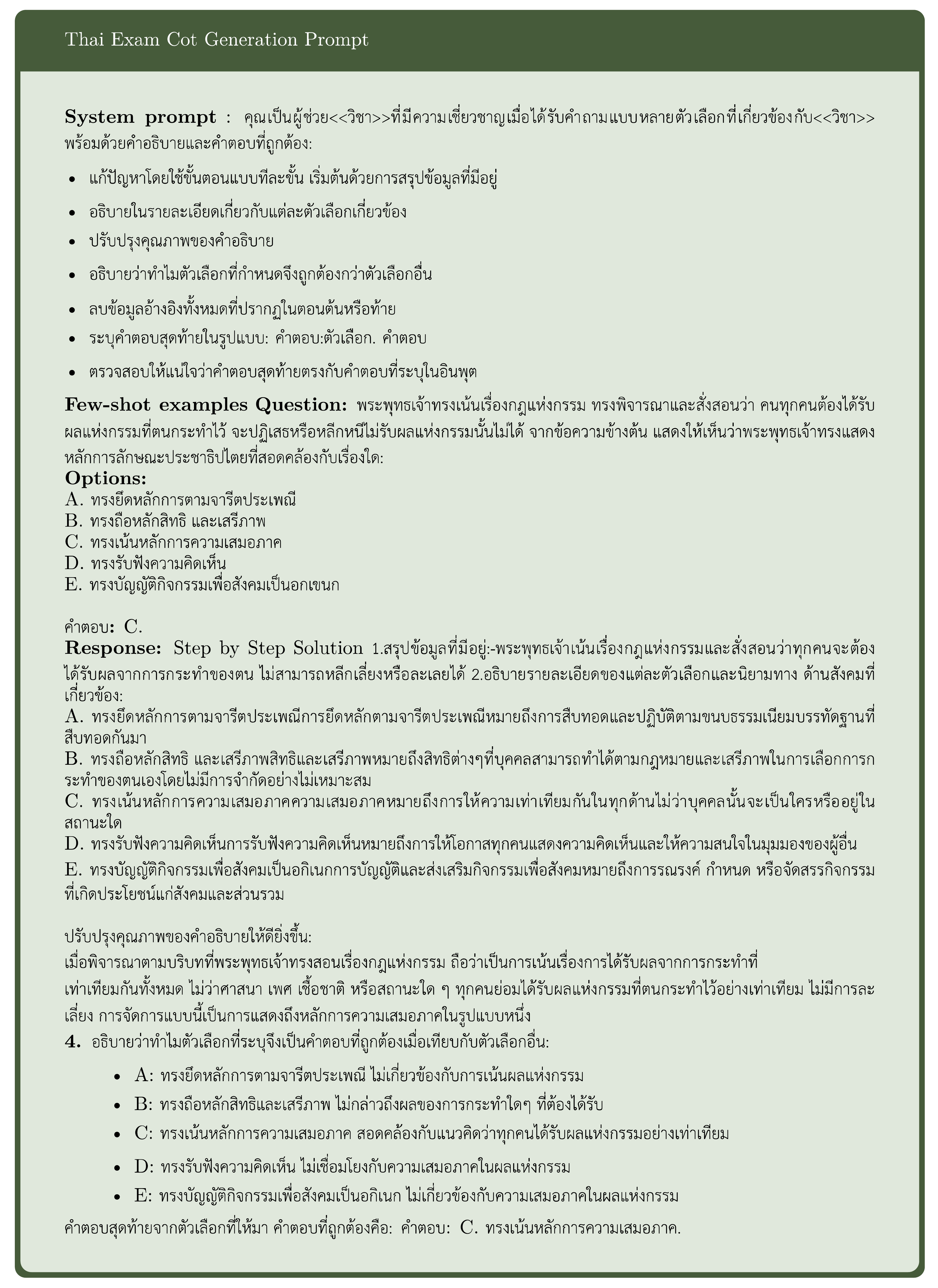}
    \label{fig:ThaiCot}
\end{figure}

\end{document}